# Robust fractional-order fast terminal sliding mode control of aerial manipulator derived from a mutable inertia parameters model

**Wenlei Zheng[1], Zhan Li[2], Bingkai Xiu[3,*], Bingliang Zhao[4], and Zhigang Guo[5]**

1. School of Astronautics, Harbin Institute of Technology, Harbin 150001, China
2. State Grid Heilongjiang Electric Power Company Limited Extrahigh Voltage Company, Harbin 150090, China[a]

Email: zhanli@hit.edu.cn.

**Abstract.** The coupling disturbance between the manipulator and the unmanned aerial vehicle (UAV) deteriorates the control performance of system. To get high performance of the aerial manipulator, a robust fractional order fast terminal sliding mode control (FOFTSMC) strategy based on mutable inertia parameters is proposed in this paper. First, the dynamics of aerial manipulator with consideration of the coupling disturbance is derived by utilizing mutable inertia parameters. Then, based on the dynamic model, a robust FOFTSMC algorithm is designed to make the system fly steadily under coupling disturbance. Furthermore, stability analysis is conducted to prove the convergence of tracking errors. Finally, comparative simulation results are given to show the validity and superiority of the proposed scheme.

## 1. Introduction

Nowadays, much attention has been paid to aerial manipulator, which is typically composed of an UAV and a multijoint manipulator. Combining the advantages of UAV and manipulator, aerial manipulator is scalable to many application scenarios, e.g., assembly, load transportation, autonomous water sampling and surveillance [1]. However, the coupling between UAV and manipulator causes the system to perform less well or even out of control. Rejection of disturbance can be divided into two steps: the first is the acquisition of disturbance and the second is the design of disturbance elimination control algorithm.

For the first step, great progress in representing coupling disturbance has been made up to now [2-4]. In [2], a disturbance observer was designed to estimate interaction force/torque. The coupling effect was obtained by a measuring device fitted between UAV and manipulator in [3]. In [4], a novel method called mutable inertia parameters was proposed to represent the disturbance. However, in [2] and [3], the mechanism of disturbance generation is not considered, and in [4], the method is only applicable in restricted situation where the motion of manipulator is slow.

For the second step, numerous control algorithms have been used to reject disturbance widely including sliding mode control (SMC) [5], adaptive control [6], impedance control [7] and intelligent control [8]. However, the above methods are all integer-order, which is challenging to integrate the theory model with experimental result effectively compared with the fractional order (FO) calculus. To date, FO has been applied to many systems in the control field [9-13]. In [9,10], two fractional order SMC methods for UAV were proposed to improve the robustness against perturbations. FO controllers were also designed to enhance the tracking performance of robotic manipulators in [11,12]. More research advances in fractional order control have been reviewed in [13]. Although fractional order technique has a wide range of applications, it is rarely seen in the control field of aerial manipulator.

Aiming at getting better performance under coupling disturbance, a robust FOFTSMC scheme based on mutable inertia parameters is proposed in this article. The main contributions are as follows.

1) A novel dynamic model derived from mutable inertia parameters is deduced to reflect the coupling disturbance. Compared with literature [2-4], this model takes full account of the disturbance generation mechanism and is suitable to fast motion scenarios.

2) The control algorithm combining fractional order with fast terminal SMC (FTSMC) is firstly proposed to reject the coupling disturbance and achieves high performance.

3) The efficiency and superiority are demonstrated by comparison with the PID controller and the FTSMC.

The rest of this paper is arranged as follows. In section 2, the dynamics of aerial manipulator based on mutable inertia parameters is presented. The control algorithm is proposed in section 3. Then, in section 4, simulation results are shown to validate the effectiveness and superiority of the proposed method. Finally, conclusions are drawn in section 5.

## 2. Dynamic model of aerial manipulator

### 2.1. Dynamics of the aerial manipulator

The model of aerial manipulator is shown in figure 1. Let $\Sigma_I$, $\Sigma_B$ and $\Sigma_i$ represent the inertia, the UAV body and the link coordinate frame, respectively. The link number is denoted by the subscript $i$. The mutable $\eta = [p^T, \Phi^T]^T$ consists of two parts: the position $p = [x, y, z]^T$ and the Euler angles $\Phi = [\phi, \theta, \psi]^T$ of UAV. The generalized mutable $q_i$ is the joint angle of the manipulator. The center of mass (COM) of aerial manipulator and manipulator are represented by point $c$ and $c_1$.

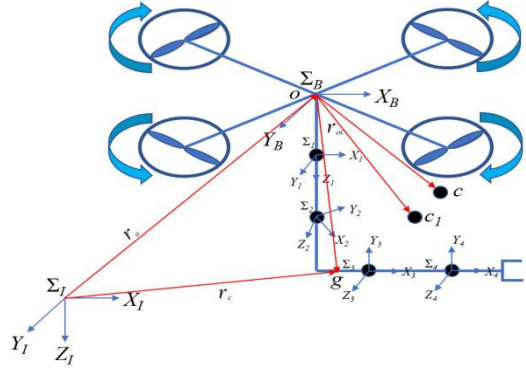

**Figure 1.** The structure of aerial manipulator

Remark1: The superscript * denotes that the mutables are expressed in the coordinate frame *.

Assuming that point g is a random mass point at any time, thus we can have where the vector $r_g$ and $r_o$ denote position of point g and point O. $r_{og}$ is the vector that the direction is $O \rightarrow g$. $^I R_B$ is the transformation matrix from $\Sigma_B$ to $\Sigma_I$.

$$r_g = r_o + r_{og} = r_o + {}^I R_B {}^B r_{og} \qquad (1)$$

Then, we can have momentum and moment of momentum of aerial manipulator as follows where $m_b$ and $m_m$ denote the mass of the UAV and the manipulator, $m_{UAM} = m_b + m_m$, $r_{oc}$ and ${}^B r_{oc1}$ are two vectors that directions are $O \to g$ and $O \to c_1$, the angular velocity of UAV body is ${}^B \omega_b$, ${}^B I_b$ and ${}^B I_m^o$ stand for the inertia matrix of the UAV and the manipulator.

$$\begin{cases} P = \int_{m_b+m_m} \dot{r}_g dm_g = m_{UAM}\dot{r}_o + m_{UAM}{}^I R_B({}^B \dot{r}_{oc1} - {}^B r_{oc1} \times {}^B \omega_b) \\ L = \int_{m_b+m_m} r_g \times \dot{r}_g dm_g = r_o \times P - m_{UAM}\dot{r}_o \times r_{oc} + {}^I R_B(({}^B I_b + {}^B I_m^o){}^B \omega_b - m_m({}^B \dot{r}_{oc1} \times {}^B r_{oc1})) \end{cases} \quad (2)$$

Based on the momentum theorem [14], differentiating (2) with respect to time yields

$$\begin{cases} \dot{P} = m_{UAM}\ddot{r}_o + m_{UAM}{}^I R_B({}^B \omega_b \times ({}^B \omega_b \times {}^B r_{oc1}) - {}^B r_{oc1} \times {}^B \dot{\omega}_b - 2{}^B \dot{r}_{oc1} \times {}^B \omega_b + {}^B \ddot{r}_{oc1}) = m_{UAM} g e_3 - F{}^I R_B e_3 \\ \dot{L} = m_{UAM}(r_{oc} \times \ddot{r}_o + \dot{r}_{oc} \times \dot{r}_o) + {}^I R_B(({}^B I_b + {}^B I_m^o){}^B \dot{\omega}_b + {}^B \dot{I}_m^{o B}\omega_b) + {}^I R_B{}^B \omega_b \times (({}^B I_b + {}^B I_m^o){}^B \omega_b) \\ + m_m{}^I R_B({}^B \omega_b \times ({}^B r_{oc1} \times {}^B \dot{r}_{oc1}) - {}^B \ddot{r}_{oc1} \times {}^B r_{oc1}) + r_o \times \dot{P} + \dot{r}_o \times P = r_o \times \dot{P} + {}^I R_B \tau + m_{UAM}{}^I R_B^B r_{oc} \times g e_3 \end{cases} \quad (3)$$

where $F$ and $\tau$ are the lift force and torque caused by the rotation of propellers, and $e_3 = [0,0,1]^T$. Therefore, the dynamics of aerial manipulator can be expressed as follows

$$\begin{cases} \dot{v}_b = -\{F{}^I R_B e_3 + m_{UAM}{}^I R_B({}^B \omega_b \times (2{}^B \dot{r}_{oc1} - {}^B r_{oc1} \times {}^B \omega_b) - {}^B r_{oc1} \times {}^B \dot{\omega}_b + {}^B \ddot{r}_{oc1})\} / m_{UAM} + g e_3 \\ ({}^B I_b + {}^B I_m^o){}^B \dot{\omega}_b = \tau + (({}^B I_b + {}^B I_m^o){}^B \omega_b) \times {}^B \omega_b + m_{UAM}({}^B r_{oc} \times ({}^B R_I g e_3 - {}^B \ddot{r}) - {}^B \dot{r}_{oc} \times {}^B \dot{r}_o) \\ -m_m({}^B \dot{r}_o \times ({}^B \dot{r}_{oc1} - {}^B r_{oc1} \times {}^B \omega_b) + {}^B \omega_b \times ({}^B r_{oc1} \times {}^B \dot{r}_{oc1}) - {}^B \ddot{r}_{oc1} \times {}^B r_{oc1}) - {}^B \dot{I}_m^{o B}\omega_b \end{cases} \quad (4)$$

Where the vector ${}^B r_{oc}$, ${}^B I_m^o$ and their derivative called mutable inertia parameters would vary during the work of manipulator, which results in the change of coupling disturbance. Thus, the disturbance can be mapped with mutable inertia parameters.

*2.2. Coupling disturbance based on mutable inertia parameters*

From (4), the coupling disturbance on the basis of mutable inertia parameters is denoted as follows

$$\begin{aligned} F_{cd} &= -m_{UAM}{}^I R_B({}^B \omega_b \times (2{}^B \dot{r}_{oc1} - {}^B r_{oc1} \times {}^B \omega_b) - {}^B r_{oc1} \times {}^B \dot{\omega}_b + {}^B \ddot{r}_{oc1}) \\ {}^B \tau_{cd} &= ({}^B I_m^{o B}\omega_b) \times {}^B \omega_b + m_{UAM}({}^B r_{oc} \times ({}^B R_I g e_3 - {}^B \ddot{r}) - {}^B \dot{r}_{oc} \times {}^B \dot{r}_o) - {}^B I_m^o {}^B \dot{\omega}_b \\ &\quad -m_m({}^B \dot{r}_o \times ({}^B \dot{r}_{oc1} - {}^B r_{oc1} \times {}^B \omega_b) + {}^B \omega_b \times ({}^B r_{oc1} \times {}^B \dot{r}_{oc1}) - {}^B \ddot{r}_{oc1} \times {}^B r_{oc1}) - {}^B \dot{I}_m^{o B}\omega_b \end{aligned} \quad (5)$$

Then, the mutable inertia parameters are calculated as

$$\begin{cases} {}^B r_{oc} = (\sum_{j=1}^n m_j {}^B p_{cj}) / m_{UAM} = (\sum_{j=1}^n m_j {}^B R_j^j p_{cj}) / m_{UAM}, {}^B r_{oc1} = m_{UAM} {}^B r_{oc} / m_m \\ {}^B \dot{r}_{oc} = (\sum_{j=1}^n m_j {}^B v_{cj}) / m_{UAM} = (\sum_{j=1}^n m_j {}^B J_{vj} \dot{q}) / m_{UAM} \\ {}^B I_m^o = \sum_{j=1}^n ({}^B R_j I_j^{cjB} R_j^{-1} + m_{j\perp}{}^B p_{cj}), {}^B \dot{I}_m^o = \sum_{j=1}^n (S({}^B \omega_j){}^B R_j I_j^{cjj} R_B - {}^B R_j I_j^{cjj} R_B S({}^B \omega_j)) + \sum_{j=1}^n m_{j\perp} p v_{cj} \end{cases} \quad (6)$$

$$\begin{cases} {}^B_\perp p_{cj} = \begin{bmatrix} {}^B p_{cjy}^2 + {}^B p_{cjz}^2 & -{}^B p_{cjx}{}^B p_{cjy} & -{}^B p_{cjx}{}^B p_{cjz} \\ * & {}^B p_{cjx}^2 + {}^B p_{cjz}^2 & -{}^B p_{cjy}{}^B p_{cjz} \\ * & * & {}^B p_{cjx}^2 + {}^B p_{cjy}^2 \end{bmatrix} \\ {}^B_\perp pv_{cj} = \begin{bmatrix} 2({}^B p_{cjy}{}^B v_{cjy} + {}^B p_{cjz}{}^B v_{cjz}) & -{}^B p_{cjx}{}^B v_{cjy} - {}^B p_{cjy}{}^B v_{cjx} & -{}^B p_{cjx}{}^B v_{cjz} - {}^B p_{cjz}{}^B v_{cjx} \\ * & 2({}^B p_{cjx}{}^B v_{cjx} + {}^B p_{cjz}{}^B v_{cjz}) & -{}^B p_{cjy}{}^B v_{cjz} - {}^B p_{cjz}{}^B v_{cjy} \\ * & * & 2({}^B p_{cjy}{}^B v_{cjy} + {}^B p_{cjx}{}^B v_{cjx}) \end{bmatrix} \end{cases} \quad (7)$$

Where ${}^B \omega_j = {}^B J_{wj} \dot{q}$, ${}^B J_{vj}$ and ${}^B J_{wj}$ represent the linear and angular velocity Jacobian matrix of *j*th link, $m_j$ denotes the mass of $j$th link. The position and velocity of $j$th link's centroid are $^Bp_{cj}$ and $^Bv_{cj}$. $I_j^{cj}$ is the inertia matrix of the $j$th link and $S(\bullet)$ is the skew symmetric matrix.

Remark2: Compared with literature [14, 15], the above makes the most of state mutables and their derivative, which can be applicable to fast motion of manipulator.

Finally, the dynamics of aerial manipulator can be simplified into.

$$\begin{cases} \dot{v}_b = (-F^I R_B e_3 + F_{cd})/m_{UAM} + ge_3 \\ ^B\dot{\omega}_b = (^BI_b)^{-1}(\tau + (^BI_b{}^B\omega_b) \times {}^B\omega_b + {}^B\tau_{cd}) \end{cases} \quad (8)$$

## 3. Controller design for aerial manipulator

In this section, a robust FOFTSMC scheme shown in figure 2 is proposed for aerial manipulator in the presence of coupling disturbance. The position and attitude loop controllers are designed in the next subsections.

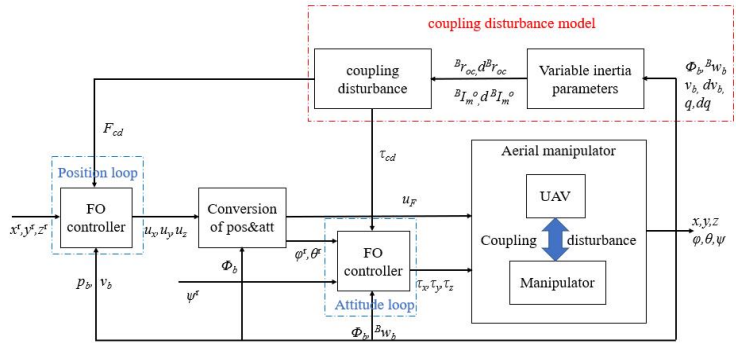

**Figure 2**. The control scheme of aerial manipulator

### 3.1. The design of position loop controller

The fractional order fast terminal sliding mode surface for the outer loop is denoted as follows where

$$S_p^{out} = \dot{e}_p + D^{\gamma_{p1}-1} c_{p1} e_p^{D_p} + e_p^{\chi} I^{\gamma_{p2}} c_{p2} e_p^{I_p}, \quad = \quad , \quad , z \quad (9)$$

$\gamma_{p1}, \gamma_{p2} \in (0,1)$, $D_p \in (1,2)$, $I_p$, $c_{p1}$ and $c_{p2}$ are all positive parameters. The tracking errors $e_p$ and $\dot{e}_p$ are given by

$$e_p = p - p^r, \quad \dot{e}_p = \dot{p} - \dot{p}^r \quad (10)$$

Remark3: $e_p^{\chi} = e_p/|e_p|$ in (9), and this representation is applicable in the following.

Differentiating (9) with respect to time leads to

$$\dot{S}_p^{out} = \ddot{e}_p + D^{\gamma_{p1}} c_{p1} e_p^{D_p} + \dot{e}_p I^{\gamma_{p2}} c_{p2} I_p |e_p|^{I_p - 1} \quad (11)$$

Letting (11) equal to 0 and consuming that coupling disturbance (5) is 0, the equivalent control is

$$u_{pc} = \ddot{p}^r - D^{\gamma_{p1}} c_{p1} e_p^{D_p} - \dot{e}_p I^{\gamma_{p2}} c_{p2} I_p |e_p|^{I_p - 1} \quad (12)$$

To achieve high performance of translational subsystem under the disturbance, we can propose the reaching control as

$$u_{pr} = -(h_{p1} + F_{cdp} + h_{p2}|S_p^{out}|)S_p^{out\chi} \quad (13)$$

where $h_{p1}, h_{p2} > 0$. Therefore, the overall control law is designed as

$$u_p = u_{pc} + u_{pr} \quad (14)$$

The stability analysis of the control scheme for the position loop is given as follows.

First, define the Lyapunov function as

$$V^{out} = \frac{1}{2} \sum_{p=x,y,z} (S_p^{out})^2 \tag{15}$$

Then, differentiating (15) with respect to time yields

$$\dot{V}^{out} = \sum_{p=x,y,z} S_p^{out} \dot{S}_p^{out} = \sum_{p=x,y,z} S_p^{out}(u_p + F_{cdp} - \ddot{p}^r + D^{\gamma_{p1}} c_p e_p^{D^p} + \dot{e}_p I^{\gamma_{p2}} c_{p2} I_p |e_p|^{l_p - 1})$$

$$= \sum_{p=x,y,z} S_p^{out} \{-(h_{p1} + F_{cdp} + h_{p2} |S_p^{out}|) S_p^{out \chi} + F_{cdp}\} \tag{16}$$

Furthermore, (16) can be represented as

$$\dot{V}^{out} \leq \sum_{p=x,y,z} \{-h_{p1}|S_p^{out}| - h_{p2}(S_p^{out})^2 - F_{cdp}|S_p^{out}| + F_{cdp}|S_p^{out}|\}$$

$$= \sum_{p=x,y,z} \{-h_{p1}|S_p^{out}| - h_{p2}(S_p^{out})^2\} = \sum_{p=x,y,z} \{-\sqrt{2} h_{p1}(V^{out})^{0.5} - 2h_{p2} V^{out}\} \tag{17}$$

Lemma1 [16]: Suppose that the Lyapunov function with initial value $V(t_0)$ is given as

$$\dot{V}(t) + \kappa_1 V(t) + \kappa_2 V^\lambda(t) \leq 0, \kappa_1, \kappa_2 > 0, 0 < \lambda < 1, V(t_0) \geq 0 \tag{18}$$

Consequently, the convergence time of finite-time stability is denoted as

$$t_s \leq t_0 + \ln\{\kappa_1 V^{1-\lambda}(t_0) / \kappa_2 + 1\} / \kappa_1 (1 - \lambda) \tag{19}$$

According to Lemma1, tracking errors of each direction $p$ will get to the sliding mode surface with a finite time

$$t_s \leq t_0 + \ln\{\sqrt{2} h_{p2} V^{0.5}(t_0) / h_{p1} + 1\} / h_{p2} \tag{20}$$

*3.2. The design of attitude loop controller*

Similar to position loop controller, the surface for the attitude loop can be described as follows:

$$S_a^{in} = \dot{e}_a + D^{\gamma_{a1}-1} c_{a1} e_a^{D_a} + e_a^{\chi} I^{\gamma_{a2}} c_{a2} e_a^{I_a}, a = \phi, \theta, \psi \tag{21}$$

where $\gamma_{a1}, \gamma_{a2} \in (0,1)$, $D_a \in (1,2)$, $I_a$, $c_{a1}$ and $c_{a2}$ are all positive parameters. The tracking errors $e_a$ and $\dot{e}_a$ are given by

$$e_a = a - a^r, \quad \dot{e}_a = \dot{a} - \dot{a}^r \tag{22}$$

Then, the attitude loop controller is designed as follows:

$$\tau_a = \{\mu_a + \ddot{\Phi}^r - D^{\gamma_{a1}} c_{a1} e_a^{D^a} - \dot{e}_a I^{\gamma_{a2}} c_{a2} I_a |e_a|^{I_a - 1} - (h_{a1} + {}^B \tau_{cdp} + h_{a2} |S_a^{in}|) S_a^{in \chi}\} / \lambda_a \tag{23}$$

where $h_{a1}$ and $h_{a1}$ are positive parameters.

The stability analysis of the control scheme for the attitude loop is given as follows.

Closely resembled the previous part, the Lyapunov is chosen as

$$V^{in} = \frac{1}{2} \sum_{p=\phi,\theta,\psi} (S_a^{in})^2 \tag{24}$$

The differentiation of (24) can be deduced as

$$\dot{V}^{in} \leq \sum_{a=\phi,\theta,\psi} \{-\sqrt{2} h_{a1} (V^{in})^{0.5} - 2h_{a2} V^{in}\} \tag{25}$$

According to Lemma1, the stability for the position and attitude loop is obtained.

## 4. Simulations

To verify the validity and superiority of the proposed control scheme, comprehensive comparisons on the PID control, FTSMC and proposed strategy are conducted.

*4.1. Simulation design*

The parameters of the UAV are set as: $m_b$ =2.65kg, $J_\phi = J_\theta = J_\psi$ =0.05kg.m², $l$ (wheelbase)=0.55m. For the 4-DOF manipulator, the mass parameters are: $m_1$ =0.238kg, $m_2$ =0.123 kg, $m_3$ =0.118kg, $m_4$ =0.224kg, the rest are listed in table 1, and the modified DH parameters are shown in table 2. The parameters of PID are considered as $kp_{x,y} = kp_{vz} = kp_{\phi,\theta}$ =4.5, $kp_z = kp_\psi$ =4, $kp_{vx,vy} = kp_r$ =2, $ki_{vx,vy} = ki_{vz} = kd_{p,q} = kd_r$ =0.03, $kd_{vz} = kd_{vx,vy} = kp_{p,q}$ =1, $ki_{p,q} = ki_r$ =0.3; the FTSMC parameters [17] are adopted; the fractional order parameters of proposed controller are $\gamma_{p1}$ =0.4, $\gamma_{a1}$ =0.2, $\gamma_{p2}$ =0.6, $\gamma_{a2}$ =0.8.

**Table 1.** Manipulator parameters

| $r$ (×10⁻³m) | | | | $I$ (×10⁻⁶kg.m²) | | | |
|---|---|---|---|---|---|---|---|
| $^1r_{c1}$ | $^2r_{c2}$ | $^3r_{c3}$ | $^4r_{c4}$ | $I^{c1}_1$ | $I^{c2}_2$ | $I^{c3}_3$ | $I^{c4}_4$ |
| $\begin{pmatrix}-6.8\\0.3\\-48.8\end{pmatrix}$ | $\begin{pmatrix}107.1\\-10.6\\0.5\end{pmatrix}$ | $\begin{pmatrix}94.3\\0\\0.5\end{pmatrix}$ | $\begin{pmatrix}60.5\\6.1\\0\end{pmatrix}$ | $\begin{bmatrix}290.2&0.3&32.5\\0.3&324.2&2.1\\32.5&2.1&141.3\end{bmatrix}$ | $\begin{bmatrix}290.2&0.3&32.5\\0.3&324.2&2.1\\32.5&2.1&141.3\end{bmatrix}$ | $\begin{bmatrix}290.2&0.3&32.5\\0.3&324.2&2.1\\32.5&2.1&141.3\end{bmatrix}$ | $\begin{bmatrix}290.2&0.3&32.5\\0.3&324.2&2.1\\32.5&2.1&141.3\end{bmatrix}$ |

**Table 2.** Modified DH parameters

| $i$ | $\alpha_{i-1}$ | $a_{i-1}$ | $d_i$ | $\theta_i$ |
|---|---|---|---|---|
| 1 | 0 | 0.012 | 0.0935 | $\theta_1$ |
| 2 | -90° | 0 | 0 | $\theta_2$-1.3855 |
| 3 | 0 | 0.13023 | 0 | $\theta_3$+1.3855 |
| 4 | 0 | 0.124 | 0 | $\theta_4$ |

During the simulation, the aerial manipulator takes off from the origin of the inertia coordinate frame and maintains hovering when it reaches an altitude of 1m. After stable hovering of system, the trajectory of each joint is set as the following at 10s.

$$q_1 = \begin{cases} 0 & ,t<10 \\ \frac{\pi}{5}\sin(\frac{\pi}{15}(t-10)) & ,t\geq 10 \end{cases}, q_2 = \begin{cases} 0 & ,t<10 \\ \frac{\pi}{3}\sin(\frac{\pi}{10}(t-10)) & ,t\geq 10 \end{cases}, q_3 = -\frac{\pi}{2}, q_4 = 0 \quad (26)$$

*4.2. Simulation results*

The simulation results are shown in figure 3, in which black line represents the desired trajectory in the absence of disturbance, the remaining lines denote real trajectories, where red one is for PID, blue one is for SMC and pink one is for our controller. The position and attitude response of system are shown in figure 3(a)-(c) and figure.3(d)-(f), which demonstrate that the proposed scheme has achieved better performance compared with PID and FTSMC.

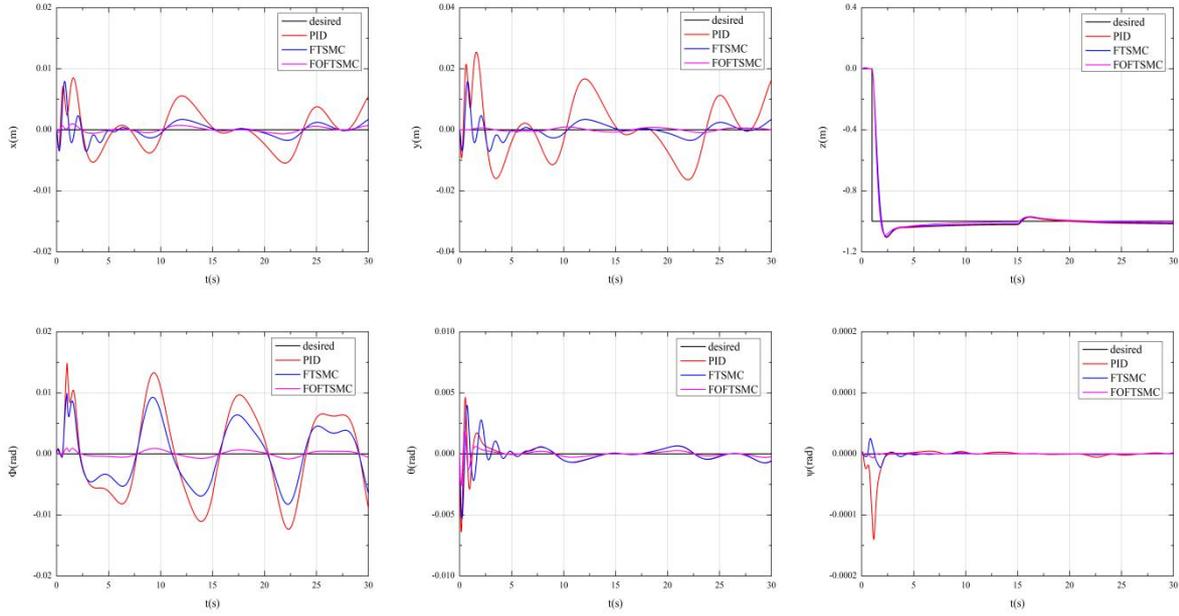

**Figure 3.** Simulation results

To further verify the superiority of the method in this article, maximum error and root mean squared error are listed in table 3.

**Table 3.** Error analysis

|  | $x$ (×$10^{-3}$) | $y$ (×$10^{-3}$) | $z$ | $\Phi$ (×$10^{-3}$) | $\psi$ (×$10^{-5}$) | $\theta$ (×$10^{-3}$) |
| --- | --- | --- | --- | --- | --- | --- |
|  | max root | max root | max root | max root | max root | max root |
| PID | 8.46 / 3.18 | 25.38 / 9.54 | 1.006 / 1.040 | 14.90 / 7.12 | -14.02 / 1.65 | -6.40 / 0.78 |
| FTSMC | 7.91 / 1.36 | 15.82 / 2.73 | 1.006 / 1.035 | 9.89 / 4.74 | -2.20 / 0.39 | -5.30 / 0.81 |
| FOFTSMC | 0.97 / 0.39 | 0.90 / 0.51 | 1.005 / 1.030 | 0.95 / 0.48 | -0.067 / 0.0079 | -2.60 / 0.31 |

## 5. Conclusion

In this article, a novel dynamic model for aerial manipulator incorporating coupling disturbance represented by mutable inertia parameters was proposed. In addition, derived from the model, a robust FOFTSMC scheme was designed to ensure that UAV maintained stable flight when the manipulator was in motion. Simulation results were shown to verify the effectiveness and superiority of the proposed method in rejecting coupling disturbance.

In the future work, experiments will be conducted to verify the validity of proposed scheme.

**Acknowledgements**

This work was supported by State Grid Heilongjiang Power Company Limited Science and Technology Project Funding.